\documentclass[11pt]{article}

\usepackage[preprint]{acl}
\usepackage{enumitem}
\usepackage{times}
\usepackage{latexsym}

\usepackage[T1]{fontenc}

\usepackage[utf8]{inputenc}

\usepackage{microtype}
\usepackage{xspace}
\usepackage{booktabs}
\usepackage{amsmath}
\usepackage{amssymb}
\usepackage{inconsolata}

\usepackage{graphicx}
\usepackage{multirow}
\usepackage{arydshln}   
\usepackage{amssymb}    
\usepackage{amsmath}    

\title{Glyph: Scaling Context Windows via Visual-Text Compression}

\author{
Jiale Cheng$^{1,2}$\thanks{\ \ Core contributors.} , Yusen Liu$^{2}$\footnotemark[1] , Xinyu Zhang$^2$\footnotemark[1] , Yulin Fei$^2$\footnotemark[1] , Wenyi Hong$^{2,3}$ \\
\textbf{Ruiliang Lyu$^2$ , Weihan Wang$^2$ , Zhe Su$^2$ , Xiaotao Gu$^2$ , Xiao Liu$^{2,3}$ , Yushi Bai$^{2,3}$}
\\
\textbf{Jie Tang$^3$, Hongning Wang$^1$  , Minlie Huang$^1$}\thanks{\ \ Corresponding author.} \\
$^1$The Conversational Artificial Intelligence (CoAI) Group, Tsinghua University \\$^2$Zhipu AI\\
$^3$The Knowledge Engineering Group (KEG), Tsinghua University\\
\small{\texttt{{chengjl23@mails.tsinghua.edu.cn,}}}  \small{\texttt{{aihuang@tsinghua.edu.cn}}}
\\
}

\begin{document}

\newcommand{\model}[0]{Glyph\xspace}

\maketitle

\begin{abstract}

Large language models (LLMs) increasingly rely on long-context modeling for tasks such as document understanding, code analysis, and multi-step reasoning.
However, scaling context windows to the million-token level brings prohibitive computational and memory costs, limiting the practicality of long-context LLMs.
In this work, we take a different perspective—visual context scaling—to tackle this challenge.
Instead of extending token-based sequences, we propose \model, a framework that renders long texts into images and processes them with vision–language models (VLMs). 
This approach substantially compresses textual input while preserving semantic information, and we further design an LLM-driven genetic search to identify optimal visual rendering configurations for balancing accuracy and compression.
Through extensive experiments, we demonstrate that our method achieves 3–4× token compression while maintaining accuracy comparable to leading LLMs such as Qwen3-8B on various long-context benchmarks.
This compression also leads to around 4× faster prefilling and decoding, and approximately 2× faster SFT training.
Furthermore, under extreme compression, a 128K-context VLM could scale to handle 1M-token-level text tasks.
In addition, the rendered text data benefits real-world multimodal tasks, such as document understanding.
Our code and model are released at \url{https://github.com/thu-coai/Glyph}.
\end{abstract}

\section{Introduction}
    

Recent advances in large language models (LLMs) have enabled remarkable progress across a wide spectrum of real-world tasks~\cite{GPT3,chowdhery2022palm,touvron2023llama,glm2024chatglm, yang2025qwen3}. 
As LLMs become increasingly capable, the demand for long-context modeling has grown critical, especially for applications such as document understanding, code analysis, and multi-hop reasoning~\cite{bai2024longbench, comanici2025gemini}. 
However, scaling context windows to hundreds of thousands or even millions of tokens poses prohibitive training and inference costs in both computation and memory, severely limiting the practicality of such models in real-world applications.


\begin{figure}[t]
    \centering
\includegraphics[width=\linewidth]{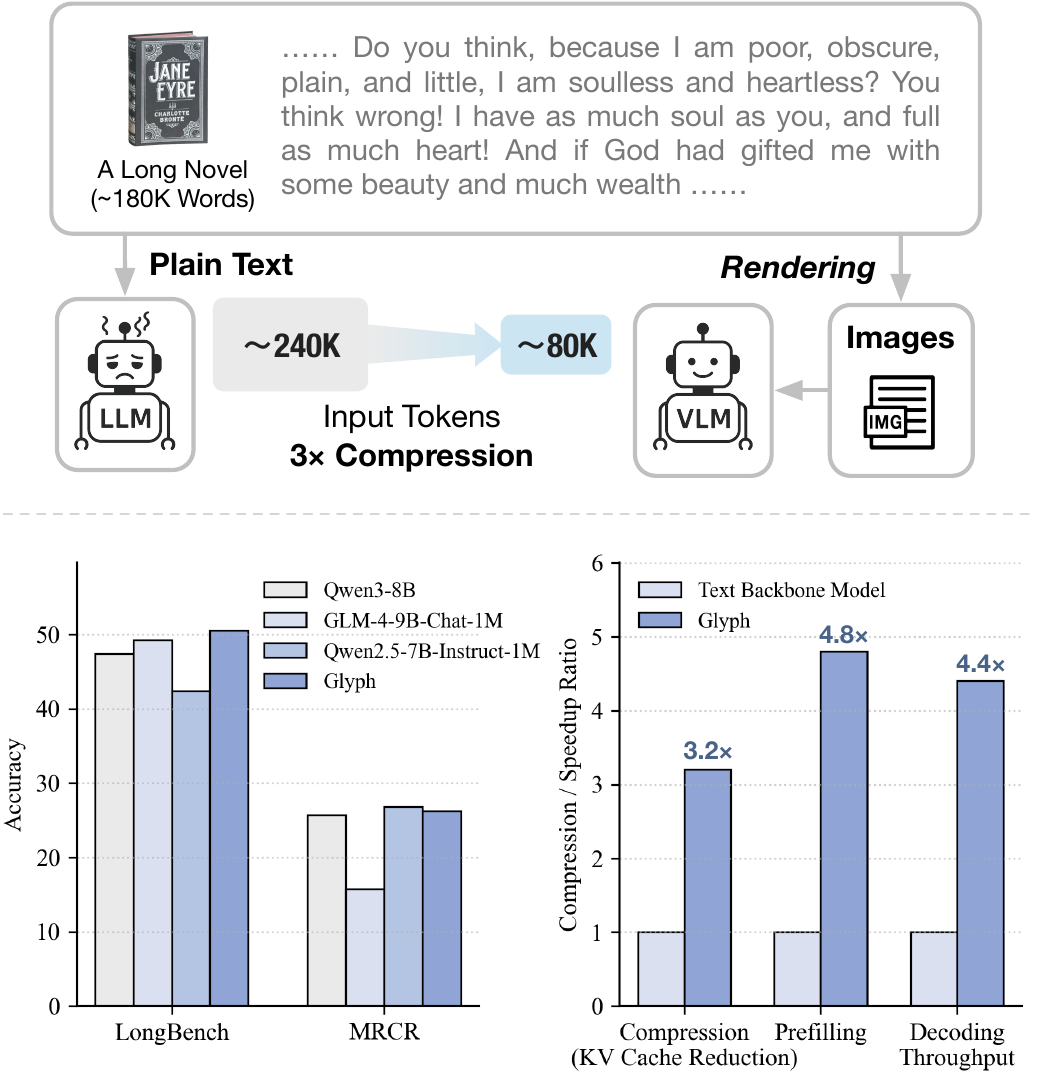}
\caption{(Upper) Comparison of two paradigms for long-context tasks: conventional approaches directly feeding plain text into LLMs, and the proposed VLM-based paradigm, Glyph, which renders text as compact images to achieve substantial input-token compression. (Lower) Glyph attains competitive performance on LongBench and MRCR, while offering significant compression and inference speedup over its text backbone model on 128K-token inputs.}
\label{fig:intro}
\vspace{-5mm}
\end{figure}


Recent work has explored two main directions to alleviate these costs. 
One line of work extends positional encodings, such as YaRN~\cite{yarn}, allowing well-trained models to accept longer inputs without additional training. 
However, such methods neither accelerate inference nor maintain accuracy when extrapolated to much longer sequences \cite{wuextending}.
Another line focuses on modifying the attention mechanism, e.g., sparse or linear attention~\cite{huang2023advancing,yang2024gated, peng2025rwkv,chen2025minimax}, which reduces the quadratic complexity of self-attention and improves per-token efficiency. 
Yet, as context length grows to hundreds of thousands of tokens, the overall overhead remains substantial, since the number of tokens is unchanged. 
Retrieval-augmented approaches \cite{laban2024summary,yu2025memagent} instead shorten the input length through external retrieval, but they risk missing important information and could introduce additional latency.


Distinct from the aforementioned approaches, we propose \model, a new paradigm that scales context length by rendering plain text into compact images and leveraging vision-language models (VLMs) to process the rendered inputs.
In this way, the VLM operates directly on the glyphs of the text—treating each visual token as a compact carrier of multiple textual tokens—thereby increasing the information density without sacrificing semantic fidelity.
This glyph-based visual representation allows a fixed-context VLM to process substantially longer texts than a text-only LLM with the same context length, thereby enabling long-context understanding without expanding the context window or relying on external retrieval mechanisms. 
For example, consider the novel ``Jane Eyre'' ($\approx$240K text tokens). A conventional 128K–context LLM cannot accommodate the entire book, and truncation easily leads to wrong answers for questions requiring global coverage, such as ``Who supports Jane when she is in distress after leaving Thornfield?'' In contrast, \model{} renders the book into compact images (e.g. $\approx$80K visual tokens), enabling a 128K–context VLM to process the full novel and answer such questions reliably.


Specifically, \model{} consists of three main stages, namely, continual pre-training, LLM-driven rendering search, and post-training.
In the continual pre-training stage, we render large-scale long-context text into diverse visual forms, enabling the VLM to transfer its long-context capability from text tokens to visual tokens. 
Since the text-to-image conversion directly determines the trade-off between context compression and model performance, devising an optimal configuration of the conversion is crucial for downstream performance.
To this end, we design an LLM-driven genetic search to automatically explore rendering parameters (e.g., font size, layout, resolution) to maximize compression while preserving long-context ability.
The resulting configuration is then applied in the post-training stage, where we perform supervised fine-tuning and reinforcement learning to further improve the model's performance on visualized input. 
An auxiliary OCR task is applied to enhance the model's ability to recognize textual content within images, thereby better aligning its visual and textual representations, yielding the final \model model.


We conduct extensive experiments to evaluate the performance of \model. Results demonstrate that \model achieves 3–4× token compression of long sequences while preserving accuracy comparable to state-of-the-art LLMs such as Qwen3-8B. 
This compression not only extends the effective context length but also improves both training and inference efficiency, yielding up to 4.8× faster prefilling, 4.4× faster decoding, as well as about 2× faster SFT training.
Moreover, we find that incorporating rendered text data effectively enhances performance on real-world multimodal long-context tasks, such as document understanding.


Our contributions can be summarized as follows:
\begin{itemize}
    \item We introduce a novel framework, \model, which enables long-context modeling through visual-text compression using VLMs, providing an alternative route to scaling context windows without incurring prohibitive computational and memory costs.
    \item We propose an LLM-driven genetic search that automatically identifies the optimal configurations of text-to-image rendering, ensuring both task performance and effective compression.
    \item We demonstrate that \model can achieve 3-4× token compression for long text sequences while preserving performance, enabling substantial improvements in memory efficiency, training, and inference speed.
\end{itemize}

\section{Related Work}

\begin{figure*}[t]
    \centering
    \includegraphics[width=\linewidth]{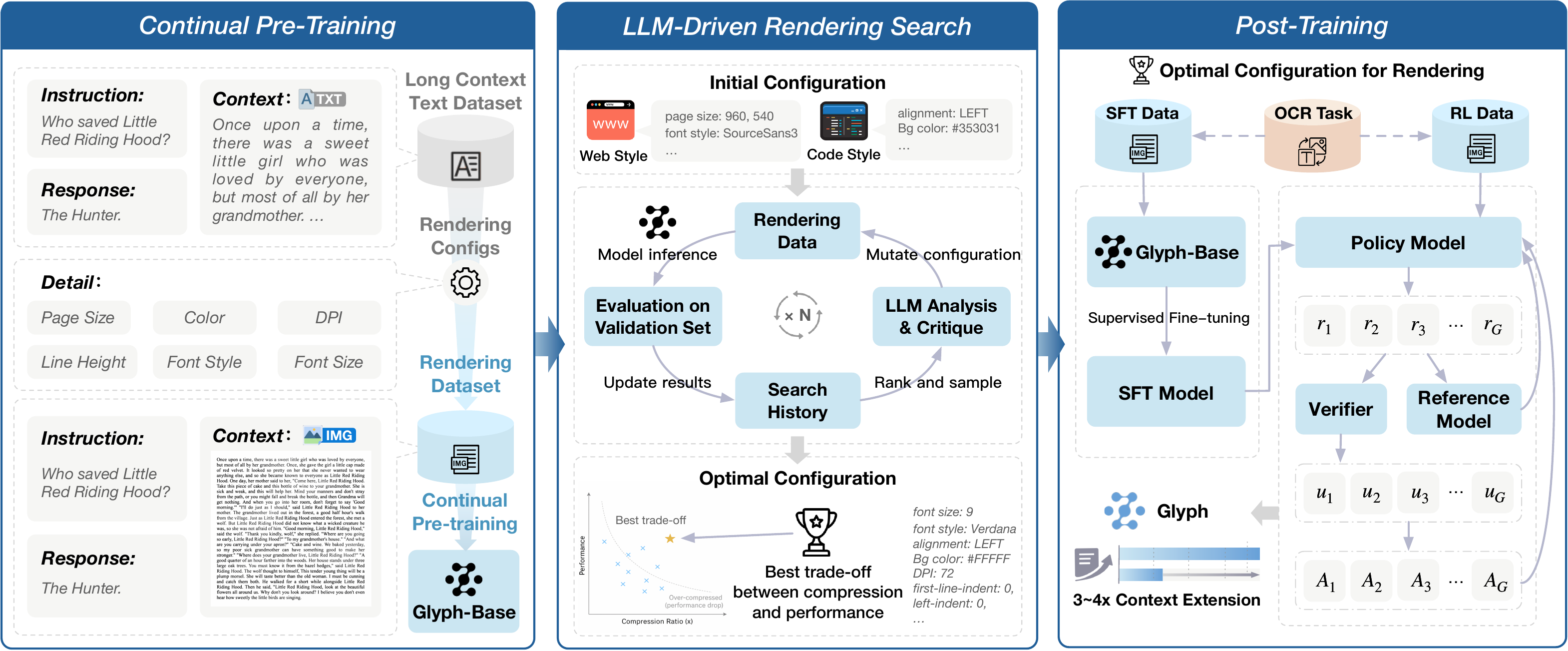}
    \caption{\model{} consists of three main stages: continual pre-training on rendered long-text data, LLM-driven genetic search for optimal rendering configurations, and post-training with SFT, RL. Together, these stages enable efficient long-context modeling with visual-text compression.}
    \label{fig:framework}
    \vspace{-5mm}
\end{figure*}

\subsection{Long-Context Modeling}
Research on extending LLMs to long contexts mainly focuses on architectural and training methods. Architecturally, studies have proposed sparse and hierarchical attention \cite{han,longformer,huang2023advancing,yang2024gated,peng2025rwkv, chen2025minimax}, positional interpolation and extrapolation \cite{rope,alibi,xpos,yarn}, and content-aware encodings \cite{cope,dape}. 
On the training side, LongAlign \cite{longalign} had built instruction datasets and loss-weighting strategies for sequences up to 100k tokens, while LongLoRA~\cite{longlora} had combined shifted sparse attention with parameter-efficient fine-tuning. LongRecipe \cite{longrecipe} had improved efficiency by integrating token analysis, index transformation, and optimization, scaling open-source models from 8k to 128k. ProLong \cite{prolong} had taken a data-centric view, selecting samples with long-range dependencies. 
In contrast, our method compresses text into visual tokens, which can be combined with existing techniques to reduce cost and extend context length.

\subsection{Multimodal Large Language Model}


Multimodal large language models (MLLMs) extend traditional LLMs to process and reason over text and visual inputs jointly. Early studies primarily focus on architectural design and effectively leveraging powerful language backbones, as exemplified by PALI~\cite{chen2022pali}, LLaVA~\cite{liu2023visual}, and CogVLM~\cite{wang2024cogvlm}.  Subsequent work further enhances these models through improvements in both LLM backbones and large-scale vision-language pretraining~\cite{hong2024cogvlm2,bai2025qwen2}, while also expanding to additional modalities such as video and audio~\cite{hurst2024gpt}.  Notably, MLLMs demonstrate strong capabilities in image perception and optical character recognition (OCR)~\cite{hong2024cogagent,liu2024llavanext}, where multiple characters or words can be represented by a single visual token, highlighting the potential for effective context compression.


\section{Method} 
    
We present \model, a novel paradigm for scaling long-context text understanding through visual compression.
Unlike conventional long-context LLMs that extend token-based context windows, \model transforms ultra-long textual inputs into compact visual images and processes them with a vision–language model.
This fundamentally different modeling method bypasses the prohibitive memory and computation costs of million-token sequences while preserving textual semantics.
Furthermore, we introduce an LLM-driven genetic search to automatically discover optimal rendering configurations, ensuring the best trade-off between compression ratio and performance.

\subsection{Overall Framework}


As illustrated in Figure~\ref{fig:framework}, \model consists of three tightly-coupled stages:
(1) Continual Pre-Training, which teaches the VLM to understand and reason over rendered long texts with diverse visual styles;
(2) LLM-Driven Rendering Search, automatically discovering the optimal rendering configuration for downstream tasks;
and (3) Post-Training, including SFT and RL under the discovered configuration to further improve the model's long-context capabilities.
Together, these stages enable \model to achieve both high accuracy and significant gains in token compression, computational efficiency, and memory usage.

\subsection{Task Definition}

\paragraph{Task Formulation.}
We formalize the standard long-context instruction following task as a triple $(\mathcal{I}, \mathcal{C}, \mathcal{R})$,
where $\mathcal{I}$ is a concise user instruction specifying the core goal,  
$\mathcal{C} = \{c_1,\dots,c_T\}$ is an ultra-long textual context,
and $\mathcal{R}$ is the target response.
The conventional learning objective is to maximize
\[
P(\mathcal{R}\mid \mathcal{I}, \mathcal{C}),
\]
i.e., to generate an accurate response conditioned on both the instruction and the long textual context.

Scaling this token-based formulation to million-token contexts, however, imposes prohibitive memory and computation costs.  
To overcome these limitations, we reformulate the input representation through \emph{visual compression}.  
Instead of directly feeding $\mathcal{C}$ as text tokens, we render it into a sequence of visual pages $\mathcal{V}=\{v_1,\dots,v_n\}$, each containing the glyphs of multiple text segments.
This allows the model to reason over a compressed but semantically equivalent input:
\[
P(\mathcal{R}\mid \mathcal{I}, \mathcal{V}).
\]
Each training instance is thus represented as $(\mathcal{I}, \mathcal{V}, \mathcal{R})$.

\paragraph{Rendering Pipeline.}
The rendering pipeline parameterizes how text is visualized before being fed into the model.  
Each rendering is specified by a configuration vector:
\[
\begin{aligned}
\boldsymbol{\theta} = \big(&
\texttt{dpi}, \texttt{page\_size}, \texttt{font\_family}, \texttt{font\_size},\\
&\texttt{line\_height}, \texttt{alignment},
\texttt{indent}, \texttt{spacing},\\
&\texttt{h\_scale}, \texttt{colors}, \texttt{borders}, \ldots\big),
\end{aligned}
\]
which controls typography, layout, and visual style of the rendered pages.
Given the context $\mathcal{C}$ and configuration $\boldsymbol{\theta}$, the pipeline produces a sequence of images that serve as the VLM's long-context input.

To quantify the degree of compression, we define the compression ratio:
\[
\rho(\boldsymbol{\theta})=\frac{|\mathcal{C}|}{\sum_{i=1}^{n}\tau(v_i)},
\]
where $\tau(v_i)$ denotes the number of visual tokens consumed by page $v_i$.
A higher $\rho$ indicates that each visual token encodes more textual information, thus achieving stronger compression.

In practice, $\boldsymbol{\theta}$ determines both information density (through font size, dpi) and visual clarity (through layout and spacing). 
By varying $\boldsymbol{\theta}$, we can continuously adjust the balance between compression and readability for the VLM.

\subsection{Continual Pre-Training}

The purpose of continual pre-training is to transfer long-context comprehension from the textual to the visual modality.
This stage exposes the VLM to a wide range of rendering styles and tasks so that it can align the semantics between rendered images and their corresponding texts.

\paragraph{Data Construction.}

To enhance model robustness, better aligning long-text capability, we adapt diverse rendering configurations over a large amount of long-context text data. 
We also develop a series of rules to exclude the improper combination of rendering parameters, e.g., a smaller line height than font size.
Moreover, with human prior, we define several style themes, including \textit{document\_style}, \textit{web\_style}, \textit{dark\_mode}, \textit{code\_style}, and \textit{artistic\_pixel}. These themes are designed to capture a wide range of document layouts and text styles, which can better exploit the knowledge that VLM has obtained in its pre-training stage.

We further introduce three families of continual pre-training tasks, including:
\begin{itemize}[leftmargin=1.5em,itemsep=0pt,parsep=0.2em,topsep=0.1em,partopsep=0.0em]
    \item \textbf{OCR Tasks}: the model reconstructs all text on one or multiple rendered pages.
    \item \textbf{Interleaved Language Modeling}: certain text spans are rendered as images, while the rest remain in text, training the model to switch seamlessly between modalities.
    \item \textbf{Generation Tasks}: given partial rendered pages (e.g., the beginning or end), the model completes the missing parts.
\end{itemize}
These tasks jointly teach the model to read, reason, and generate under visually compressed contexts.

\paragraph{Loss Function.}
We minimize the cross-entropy loss
\begin{equation}
\mathcal{L}_{\text{CPT}} 
= - \mathbb{E}_{(\mathcal{I}^{\!*}, \mathcal{V}, \mathcal{R})} 
\sum_{t} \log P_\phi(r_t \mid \mathcal{I}^{\!*}, \mathcal{V}, r_{<t}) ,
\end{equation}
where $\mathcal{I}^{\!*}$ denotes an optional instruction 
(e.g., absent in interleaved language modeling tasks)
and $\phi$ is initialized from the base VLM.
This stage produces a model capable of understanding rendered text and handling long contexts, referred to as \model-Base.

\subsection{LLM-Driven Rendering Search}


Although diverse rendering improves generalization, downstream tasks often require a specific trade-off between compression and visual clarity for the VLM.
We therefore perform an LLM-driven genetic search after continual pre-training to automatically identify the optimal rendering configuration $\boldsymbol{\theta}^{*}$ used in the post-training stage.

\paragraph{Genetic Algorithm.}
Starting from an initial population of candidate configurations $\{\boldsymbol{\theta}_k\}$ sampled from pre-training configurations, we iteratively perform the following steps:
\begin{enumerate}[leftmargin=1.5em,itemsep=0pt,parsep=0.2em,topsep=0.1em,partopsep=0.0em]
    \item \textbf{Rendering Data}: render the validation set using each configuration $\boldsymbol{\theta}_k$ to obtain visual inputs.
    \item \textbf{Evaluation on Validation Set}: perform model inference on the rendered data, measure task accuracy and compression ratio, and update the results.
    \item \textbf{LLM Analysis \& Critique}: use an LLM to suggest promising mutations and crossovers based on the current population and validation results.
    \item \textbf{Search History}: record all configurations and their performance; rank and sample promising candidates for the next iteration.
\end{enumerate}
This process continues until the population converges, i.e., when no further improvement is observed in validation accuracy or compression over a pre-defined number of generations.
The resulting configuration $\boldsymbol{\theta}^{*}$ is then adopted for post-training.

\subsection{Post-Training}



With the optimal rendering configuration $\boldsymbol{\theta}^{*}$ fixed, we further improve \model-Base through two complementary optimization stages—\emph{supervised fine-tuning} and \emph{reinforcement learning}—supplemented by an \emph{auxiliary OCR alignment} task.
Together, these components jointly enhance the model's ability to reason over visually compressed inputs and to recognize textual details.

\paragraph{Supervised Fine-Tuning.}
To endow the model with robust comprehension under visual inputs, we curate a high-quality text SFT corpus and render its long-context inputs using the optimal configuration. 
Each response adopts a thinking-style format, in which each example contains explicit reasoning traces (e.g., ``\texttt{<think>...</think>}'').  
This encourages the model to perform step-by-step reasoning when reading massive token contexts.

Formally, the loss function can be written as
\begin{equation}
\mathcal{L}_{\text{SFT}} 
= - \mathbb{E}_{(\mathcal{I}, \mathcal{V}, \mathcal{R})} 
\sum_{t} \log P_\phi(r_t \mid \mathcal{I}, \mathcal{V}, r_{<t}) ,
\end{equation}
where $\phi$ is initialized from the continual pre-training checkpoint.
This stage establishes a strong initialization for reinforcement learning.
\begin{table*}[htbp]
\centering
\resizebox{0.98\linewidth}{!}{%
\begin{tabular}{@{}l ccccccccccccc@{}}  
\toprule
\multirow{2}{*}{\textbf{Model}} & \multicolumn{2}{c}{\textbf{Single-Doc QA}} & \multicolumn{2}{c}{\textbf{Multi-Doc QA}} & \multicolumn{2}{c}{\textbf{Summarization}} & \multicolumn{2}{c}{\textbf{Few-shot}} & \multicolumn{2}{c}{\textbf{Synthetic}} & \multicolumn{2}{c}{\textbf{Code}} & \multirow{2}{*}{\textbf{Avg}} \\  
\cmidrule(lr){2-3} \cmidrule(lr){4-5} \cmidrule(lr){6-7} \cmidrule(lr){8-9} \cmidrule(lr){10-11} \cmidrule(lr){12-13}
& \textbf{QP} & \textbf{NQA} & \textbf{HQA} & \textbf{2QA} & \textbf{QSUM} & \textbf{GovRep} & \textbf{TREC} & \textbf{TriQA} & \textbf{PR Zh} & \textbf{PR En} & \textbf{RB} & \textbf{LCC} & \\
\midrule
\textcolor{gray}{GPT-4.1} & \textcolor{gray}{51.60} & \textcolor{gray}{35.73} & \textcolor{gray}{69.10} & \textcolor{gray}{74.15} & \textcolor{gray}{23.50} & \textcolor{gray}{33.36} & \textcolor{gray}{77.00} & \textcolor{gray}{93.36} & \textcolor{gray}{100.00} & \textcolor{gray}{100.00} & \textcolor{gray}{67.94} & \textcolor{gray}{68.43} & \textcolor{gray}{56.03} \\
\hdashline 
\addlinespace 
LLaMA-3.1-8B-Instruct & 44.56 & 26.34 & 56.88 & 46.67 & \textbf{23.28} & \textbf{32.36} & 19.25 & \underline{89.12} &  62.20  & \underline{99.50} & 42.81 & 46.35 & 41.34 \\
Qwen2.5-7B-Instruct-1M & \textbf{45.29} & 25.61 & 60.70 & 40.51 & \underline{22.95} & \underline{29.97} & 59.37 &  86.93 & \underline{98.5} & \textbf{100.00} & 29.80 & 21.72 & 42.42 \\
Qwen3-8B & \underline{44.67} & 26.13 & \underline{65.83} & \textbf{73.92} & 19.60 & 26.85 & \underline{70.50} & 87.98 & \textbf{100.00} & 97.26 & 40.89 & 44.87 & 47.46 \\
GLM-4-9B-Chat-1M  & 43.75 & \underline{26.72} & 58.98 & 50.89 & 22.84 & 27.60 & 61.50 & \textbf{90.07} & \textbf{100.00} & \underline{99.50} & \underline{55.64} & \textbf{59.54} & \underline{49.27} \\
\midrule
\model  & 40.64 & \textbf{28.45} & \textbf{66.42} & \underline{72.98} & 19.78 & 25.53 & \textbf{82.62} &  88.54 & 89.03 & \underline{99.50} & \textbf{60.80} & \underline{48.85} & \textbf{50.56} \\
\bottomrule
\end{tabular}%
}
\vspace{-0.1cm}
\caption{Performance comparison of \model with leading LLMs on LongBench (\%). Our model achieves competitive results in the overall average score. 
Best results are \textbf{bolded}, and second-best are \underline{underlined}.
Refer to Table \ref{tab:longbench_subtasks} for the rest of the results.}
\vspace{-2mm}
\label{table: longbench main}
\end{table*}

\begin{table*}[ht]
\centering
\resizebox{0.98\linewidth}{!}{%
\begin{tabular}{l *{12}{c}}  
\toprule
\multirow{3}{*}{\textbf{Model}} & \multicolumn{6}{c}{\textbf{4 Needle}} & \multicolumn{6}{c}{\textbf{8 Needle}} \\
\cmidrule(lr){2-7} \cmidrule(lr){8-13}
& \textbf{0k-8k} & \textbf{8k-16k} & \textbf{16k-32k} & \textbf{32k-64k} & \textbf{64k-128k} & \textbf{Avg} 
& \textbf{0k-8k} & \textbf{8k-16k} & \textbf{16k-32k} & \textbf{32k-64k} & \textbf{64k-128k} & \textbf{Avg} \\
\midrule
\textcolor{gray}{GPT-4.1} & \textcolor{gray}{50} & \textcolor{gray}{38} & \textcolor{gray}{29} & \textcolor{gray}{42} & \textcolor{gray}{38} & \textcolor{gray}{39.4}  
& \textcolor{gray}{33} & \textcolor{gray}{26} & \textcolor{gray}{17} & \textcolor{gray}{22} & \textcolor{gray}{19} & \textcolor{gray}{23.4}  \\
\hdashline 
\addlinespace 
LLaMA-3.1-8B-Instruct   & \underline{33.42} & \underline{25.97} & \underline{22.73} & \textbf{26.97} & 12.68 & \underline{24.35} & \underline{23.80} & 17.69 & \textbf{19.85} & \underline{17.72} & 11.79 & \textbf{18.17} \\
Qwen2.5-7B-Instruct-1M    & 25.96 & 20.13 & 19.93 & 24.25 & \underline{17.29} & 21.51 & 17.64 & 19.48 & 12.41 & 14.80 & \underline{14.24} & 15.71 \\
Qwen3-8B      & 29.34 & 22.67 & 20.34 & 23.63 & \textbf{19.11} & 23.02 & 18.75 & \underline{19.69} & \underline{16.81} & \textbf{17.86} & \textbf{15.00} & 17.62 \\
GLM-4-9B-Chat-1M        & 15.17 & 13.78 & 9.18 & 20.27 & 15.05 & 14.69 & 14.55 & 9.65 & 9.34 & 9.47 & 8.97 & 10.40 \\
\midrule
Glyph         & \textbf{35.44} & \textbf{26.82} & \textbf{24.15} & \underline{25.69} & 16.37 & \textbf{25.81} & \textbf{25.12} & \textbf{21.22} & 16.43 & 13.91 & 13.51 & \underline{18.14} \\
\bottomrule
\end{tabular}%
}
\vspace{-0.1cm}
\caption{Performance comparison of our model against leading LLMs on the 4-needle and 8-needle sub-tasks of the MRCR benchmark (\%). Our method consistently ranks first or second across most settings while preserving about 3$\times$ compression ratio. Performance on the 2-needle task is deferred to the Appendix.}
\label{table: mrcr main}
\vspace{-3mm}
\end{table*}

\begin{figure*}[h!]
\centering
\includegraphics[width=0.95\linewidth]{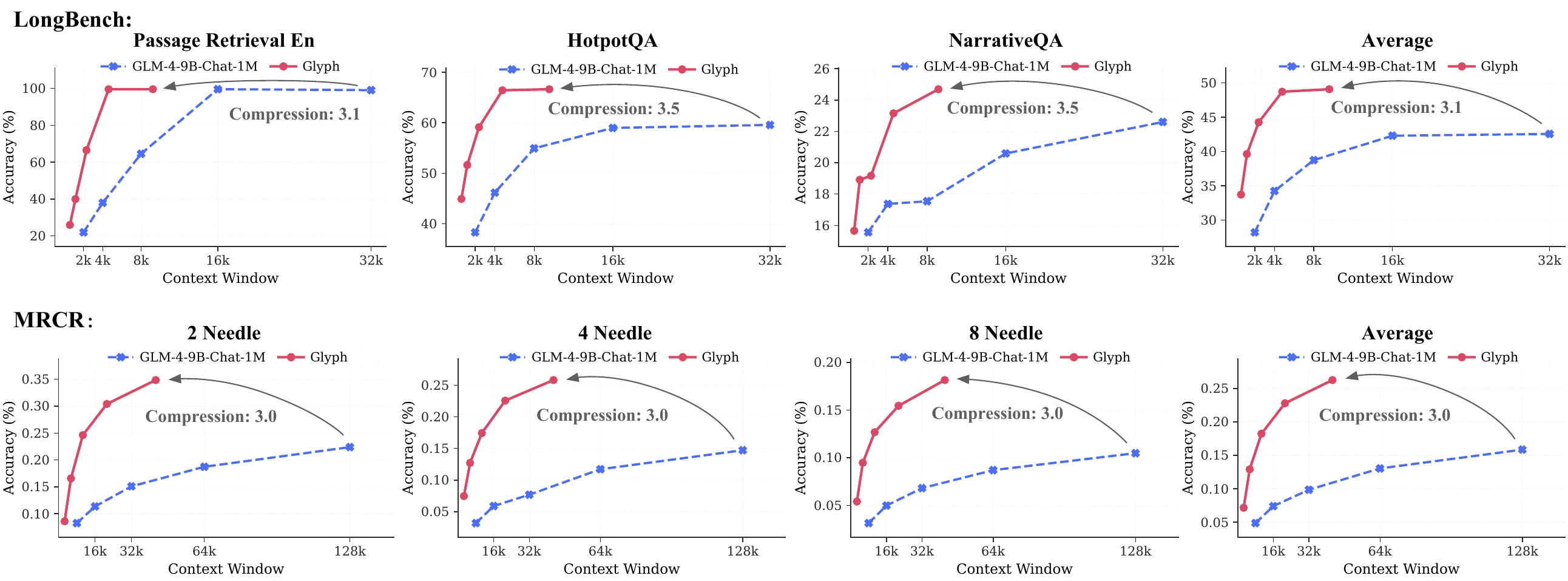}
\caption{Performance comparison of Glyph and the baseline across different context windows, demonstrating that Glyph achieves performance equivalent to longer contexts with substantially shorter context windows.}
  \vspace{-0.5cm}
   \label{fig:acc_length}
\end{figure*}

\paragraph{Reinforcement Learning.}
After SFT, we further refine the policy using Group Relative Policy Optimization (GRPO). 
For each input $(\mathcal{I}, \mathcal{V})$, we sample a group of candidate responses 
$\{r_1,\ldots,r_G\}$ from the old policy $\pi_{\phi_{\text{old}}}$. 
We first define the importance sampling weight:
\begin{equation}
w_i \;=\; \frac{\pi_\phi(r_i \mid \mathcal{I}, \mathcal{V})}{\pi_{\phi_{\text{old}}}(r_i \mid \mathcal{I}, \mathcal{V})} .
\end{equation}
Each sampled response $r_i$ receives a reward score 
$u(r_i)\in\{0,1\}$, which integrates:
\begin{itemize}
    \item \textbf{Verifiable rewards} from an external LLM judge, scoring based on the accuracy of the answer, which is a reference-based LLM-as-a-judge with the reference being the ground truth.
    \item \textbf{Format rewards} that ensure the response correctly follows the defined thinking style.
\end{itemize}  
The group-normalized advantage is computed as:
\begin{equation}
A_i \;=\; \frac{u(r_i) - \mathrm{mean}(\{u(r_j)\}_{j=1}^G)}
{\mathrm{std}(\{u(r_j)\}_{j=1}^G)} ,
\end{equation}
and the GRPO objective is
\begin{equation}
\begin{aligned}
\mathcal{J}_{\text{GRPO}}(\phi)
= \;& \mathbb{E}_{(\mathcal{I}, \mathcal{V}) \sim P,\, \{r_i\}_{i=1}^G \sim \pi_{\phi_{\text{old}}}}
\Bigg[ \frac{1}{G} \sum_{i=1}^G \Big( \\
&
\min \big( w_i A_i,\;
\mathrm{clip}(w_i,\, 1-\epsilon_l,\, 1+\epsilon_h)\, A_i \big) \\
& 
-~ \beta\, D_{\mathrm{KL}}\!\big(\pi_\phi \,\|\, \pi_{\text{SFT}}\big)
\Big) \Bigg],
\end{aligned}
\end{equation}
where $\epsilon$ and $\beta$ are hyperparameters.


\paragraph{Auxiliary OCR Alignment.}
A persistent challenge of visual compression is the faithful recovery of fine-grained text from rendered images.  
Throughout both SFT and RL, we therefore incorporate an auxiliary OCR alignment task that encourages the model to correctly read and reproduce low-level textual details.  
The form of the OCR task is the same as in the continual pre-training stage. In the RL stage, the reward for the OCR task is given by the Levenshtein distance.

By integrating structured SFT supervision, RL optimization, and continuous OCR-aware alignment, \model acquires both powerful long-context reasoning ability and stable low-level text recognition, achieving strong downstream performance under highly compressed visual contexts.


\section{Experiments}

\subsection{Experimental Setup}

To comprehensively evaluate the effectiveness of our method, we have conducted extensive experiments covering long-context understanding, efficiency, cross-modal generalization, and several ablations and analysis. Implementation details, descriptions of baselines and benchmarks are provided in Appendix \ref{appendix: implementatio}.

\subsection{Main Results on Performance}

\begin{table*}[ht]
\centering
\resizebox{0.95\linewidth}{!}{%
\begin{tabular}{lcccccccccccc}
\toprule
\textbf{Model} & \textbf{Niah-S1} & \textbf{Niah-S2} & \textbf{Niah-M1} & \textbf{Niah-M2} & \textbf{Niah-V} & \textbf{Niah-Q} & \textbf{VT} & \textbf{CWE} & \textbf{FWE} & \textbf{QA-1} & \textbf{QA-2} & \textbf{Avg} \\
\midrule
\textcolor{gray}{GPT-4.1} & \textcolor{gray}{100.0} & \textcolor{gray}{98.85} & \textcolor{gray}{100.0} & \textcolor{gray}{100.0} & \textcolor{gray}{99.67} & \textcolor{gray}{100.0} & \textcolor{gray}{100.0} & \textcolor{gray}{97.87} & \textcolor{gray}{98.66} & \textcolor{gray}{86.82} & \textcolor{gray}{77.47} & \textcolor{gray}{96.30} \\
\hdashline
\addlinespace
LLaMA-3.1-8B-Instruct & 99.33 & 99.33 & 99.33 & \textbf{99.00} & 98.17 & \underline{99.67} & 87.07 & 57.30 & 81.85 & \textbf{84.00} & 58.00 & 87.55\\ 
Qwen2.5-7B-Instruct-1M  & \textbf{100.00} & \underline{99.67} & \underline{99.67} & \textbf{99.00} & 93.83 & 98.75 & 85.40 & 72.10 & 85.67 & \underline{80.00} & 60.67 & 88.61\\
Qwen3-8B & \textbf{100.00} & \textbf{100.00} & 95.33 & 84.67 & 97.42 & 99.33 & \underline{98.47} & 74.67 & 86.67 & 70.33 & 53.33 & 87.29\\
GLM-4-9B-Chat-1M & \textbf{100.00} & \textbf{100.00} & 92.67 & \textbf{99.00} & 95.00 & \textbf{100.00} & 98.20 & 49.50 & 83.22 & 72.67 & 56.67 & 86.08\\
\midrule
\multicolumn{13}{c}{\textbf{DPI: 72 / Compression rate: average 4.0, up to 7.7}} \\
\midrule
Glyph & 73.33 & 64.67 & 67.33 & 56.00 & 73.42 & 71.42 & 77.93 & 94.40 & 92.67 & 59.33 & 63.33 & 72.17 \\
\midrule
\multicolumn{13}{c}{\textbf{DPI: 96 / Compression rate: average 2.2, up to 4.4}} \\
\midrule
Glyph & 98.00 & 95.33 & 95.67 & 85.00 & 96.33 & 95.83 & 94.93 & \underline{94.80} & \underline{98.00} & 79.00 & \underline{70.67} & \underline{91.23} \\
\midrule
\multicolumn{13}{c}{\textbf{DPI: 120 / Compression rate: average 1.2, up to 2.8}} \\
\midrule
Glyph & \underline{99.67} & 99.00 & \textbf{100.00} & \underline{93.67} & \textbf{99.00} & 99.58 & \textbf{99.33} & \textbf{98.97} & \textbf{99.11} & 79.00 & \textbf{74.00} & \textbf{94.67} \\
\bottomrule
\end{tabular}%
}
\vspace{-0.1cm}
\caption{Performance on the Ruler benchmark (\%). We demonstrate the impact of different DPI settings on our model's performance and the resulting compression ratios. For each configuration, the table includes both the average compression ratio across all sub-tasks and the maximum compression achieved for specific sub-task types.}
\label{tab:ruler_single_header}
\end{table*}

\begin{figure*}[ht]
\centering
\includegraphics[width=1\linewidth]{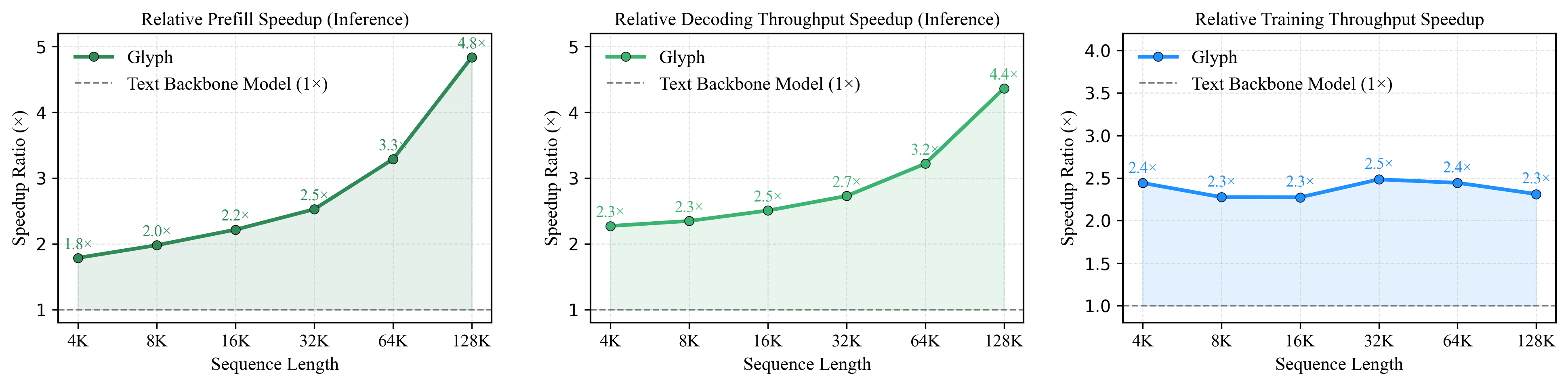}
\vspace{-0.8cm}
\caption{Speedup ratios of Glyph over the text backbone model for prefill, decoding, and training across different sequence lengths.}
  \vspace{-0.3cm}
   \label{fig: speedup}
   
\end{figure*}

\begin{figure}
    \centering
    \includegraphics[width=0.95\linewidth]{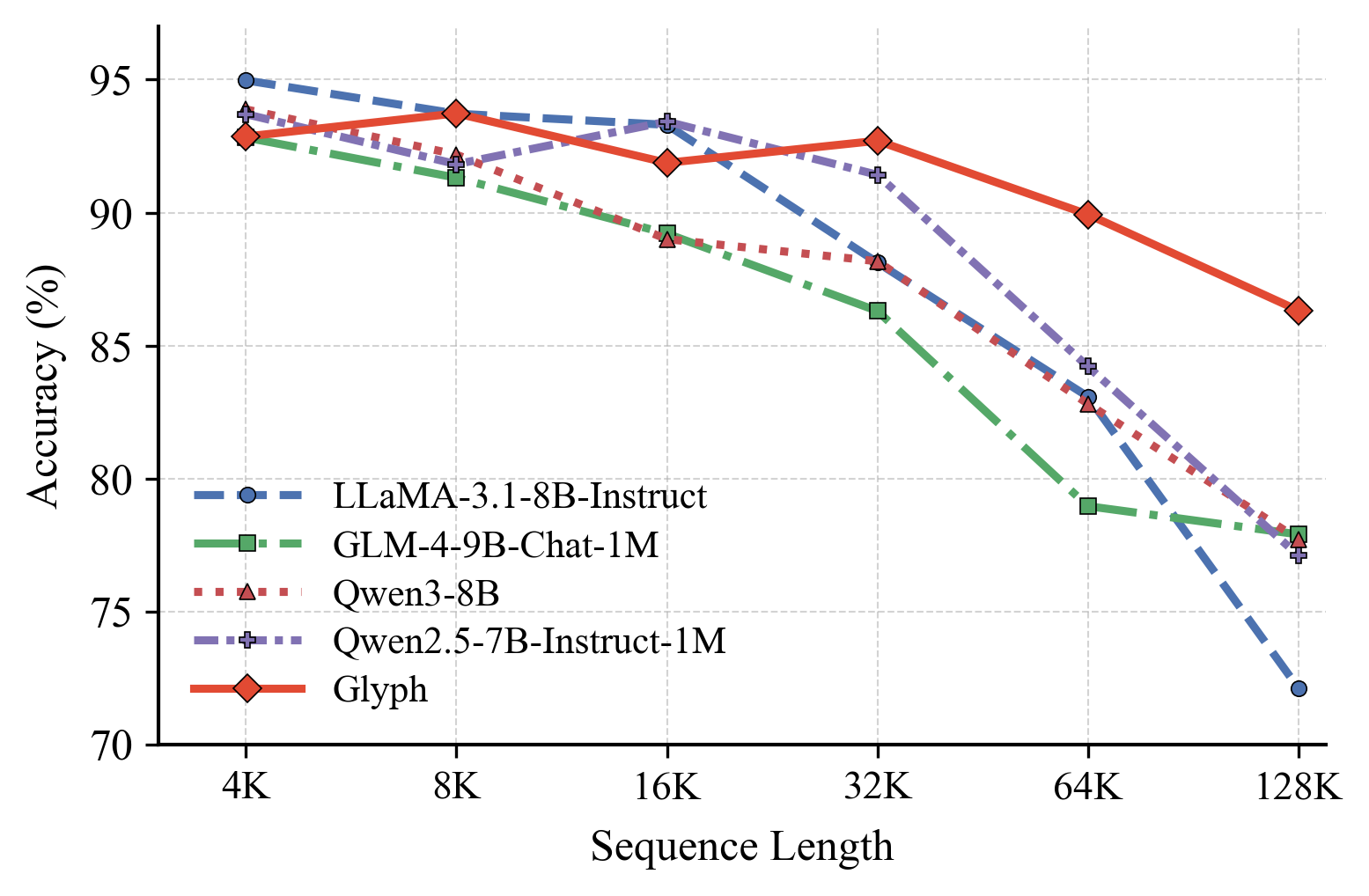}
    \vspace{-4mm}
    \caption{Model performance degradation across different sequence lengths on the Ruler benchmark.}
    \vspace{-5mm}
    \label{fig:ruler length}
\end{figure}

\subsubsection{Results on LongBench \& MRCR}


Tables~\ref{table: longbench main} and \ref{table: mrcr main} summarize overall results. 
\model achieves performance on par with or surpassing state-of-the-art text-only 
LLMs of similar size, including Qwen3-8B and GLM-4-9B-Chat-1M, demonstrating that \model remains effective on long-context tasks with a large reduction in input tokens.

Figure \ref{fig:acc_length} further illustrates the context scaling behavior of \model. For LongBench, we report the results with truncated contexts; for MRCR, we utilize the original dataset split. 
On LongBench, our model achieves an average effective compression ratio of 3.3$\times$, with certain tasks reaching up to around 5$\times$. 
On MRCR, the average compression ratio is 3.0$\times$. 
This means that within the same token budget, \model can effectively 
utilize several times more original context than text-only models.  
More importantly, as the input length grows, this advantage scales up.
When a text-only model extends its window from 32k to 64k tokens, it gains 32k additional tokens of usable context. 
Under the same expansion, \model—with a compression ratio of around 3$\times$—effectively gains about 96k tokens' worth of original text. 
This advantage translates into a faster improvement as the context length increases.

\subsubsection{Results on Ruler}



On the Ruler benchmark, \model also achieves performance comparable to leading LLMs across most categories (Table \ref{tab:ruler_single_header}).
We exclude the UUID task from this benchmark due to its huge difficulty for VLMs, which is further discussed in the limitations section.  

Beyond raw scores, we demonstrate the advantage of test-time scaling. 
When we increase the rendering resolution (DPI) at inference time, our model shows 
substantial gains: at higher DPI settings, it even surpasses strong text-only baselines. 
This demonstrates that the performance of VLMs on text-only long-context tasks has a high ceiling, and that \model still holds considerable potential.

Furthermore, we analyze performance under different sequence lengths 
(Figure~\ref{fig:ruler length}). At short contexts, text-only models such as LLaMA-3.1-8B-Instruct maintain a slight edge. 
However, as the input length grows, \model exhibits obviously slower degradation. 
This aligns with the earlier observations on LongBench and MRCR: thanks to compression, an increase in the nominal text context window translates to a much smaller increase in the effective length the \model model actually needs to handle. Consequently, our model 
maintains accuracy more stably as the context grows.

\begin{table}[t]
\centering
\renewcommand{\arraystretch}{1.25}
\resizebox{\linewidth}{!}{
\begin{tabular}{lccccc}
\toprule
\textbf{Model} & \textbf{SP} & \textbf{CP} & \textbf{UA} & \textbf{Acc} & \textbf{F1} \\
\midrule
GLM-4.1V-9B-Base     & 36.76 & \underline{23.41} & \underline{21.52} & 29.18 & 28.78 \\ \midrule
\model-Base   & \underline{47.91} & 22.24 & 14.80 & \underline{32.48} & \underline{34.44} \\
\model     & \textbf{57.73} & \textbf{39.75} & \textbf{27.80} & \textbf{45.57} & \textbf{46.32} \\
\bottomrule
\end{tabular}
}
\caption{Results on MMLongBench-Doc (\%). 
SP, CP, UA, and Acc denote Single-page, Cross-page, Unanswerable, and Overall Accuracy, respectively.}
\label{tab:mmlongbench-doc}
\vspace{-3mm}
\end{table}

\begin{table}[t]
\centering
\resizebox{\linewidth}{!}{
\begin{tabular}{lcccc}
\toprule
\textbf{Configuration} & \textbf{LongBench} & \textbf{MRCR} & \textbf{Ruler} & \textbf{Avg.} \\
\midrule
Random Config  & \underline{41.78} & 15.82 & 65.13 & 40.91 \\
Manual Config  & \textbf{43.45} & \underline{19.33} & \underline{68.09} & \underline{43.62} \\ \midrule
Search-based Config & \textbf{43.45} & \textbf{22.10} & \textbf{71.24} & \textbf{45.60} \\
\bottomrule
\end{tabular}}
\caption{Ablation study comparing randomly combined, manually designed, and search-based configurations on three benchmarks under SFT setting. The search-based configuration achieves the best overall performance.}
\label{tab:ablation_configs}
\vspace{-3mm}
\end{table}

\begin{table}[t]
\centering
\renewcommand{\arraystretch}{1.15}
\setlength{\tabcolsep}{8pt}
\resizebox{\linewidth}{!}{
\begin{tabular}{lccc}
\toprule
\textbf{Model} & \textbf{LongBench} & \textbf{MRCR} & \textbf{Ruler} \\
\midrule
\model              & 50.56 & 26.27 & 72.17 \\
\midrule
-- w/o OCR (in RL)        & -1.40 & -2.00 & -0.35 \\
-- w/o RL                 & -7.11 & -4.17 & -0.93 \\
-- w/o OCR (in SFT)       & -8.12 & -8.42 & -1.23 \\
\bottomrule
\end{tabular}
}
\caption{Ablation study showing the performance drop (\%) relative to the final \model model when components are progressively removed.}
\label{tab:ablation_relative}
\vspace{-2mm}
\end{table}

\begin{table}[ht]
\resizebox{\linewidth}{!}{
\centering
\begin{tabular}{lccc}
\toprule
\textbf{Model} & \textbf{2 Needle} & \textbf{4 Needle} & \textbf{8 Needle} \\
\midrule
GLM-4-9B-Chat-1M       & \underline{10.08} & 6.19 & 2.26 \\
Qwen2.5-7B-Instruct-1M & \textbf{11.36} & \underline{7.34} & \textbf{7.77} \\ \midrule
\model           & 9.36 & \textbf{7.62} & \underline{7.64} \\
\bottomrule
\end{tabular}}
\caption{
Average MRCR performance (\%) across 128K–1M context lengths under different needle counts.
}
\label{tab:mrcr-128k}
\end{table}

\subsection{Efficiency Evaluation}



We further evaluate the efficiency of our method in both training and inference, comparing \model with the text backbone model. 
The evaluation setting is detailed in Appendix \ref{appendix: implementatio}.
As shown in Figure~\ref{fig: speedup}, \model provides clear speedups in both metrics, demonstrating significant gains at the inference stage and SFT training stage.
As the sequence length grows from 8k to 128k, our model demonstrates markedly better scalability, achieving stable SFT training throughput speedup and growing inference speedup.

\subsection{Cross-Modal Generalization}

Although our training data mainly consists of rendered text images rather than natural multimodal inputs, we are interested in whether such training can generalize to real-world multimodal tasks, like long document understanding. 
To this end, we evaluate \model on the MMLongBench-Doc benchmark, which contains 130 long PDF documents with diverse layouts and embedded images. 
As shown in Table~\ref{tab:mmlongbench-doc}, \model achieves clear improvements over our backbone model GLM-4.1V-9B-Base, confirming its ability to generalize across modalities.

\subsection{Ablation Study \& Analysis}




We conduct a series of ablations and analyses to better understand our method.

\paragraph{Configuration Search.} 
We compare three types of rendering configurations for SFT: (i) randomly sampled configuration from the pre-training sets, (ii) manually designed settings based on prior knowledge, and (iii) the configuration obtained from our search procedure.
While all settings achieve comparable compression ratios, Table~\ref{tab:ablation_configs} shows that the searched configuration consistently outperforms the other two, both on average and across most individual tasks.
This demonstrates the importance of systematic exploration for finding appropriate rendering strategies.

\paragraph{OCR Auxiliary Tasks.} 
We also test the impact of adding OCR auxiliary tasks during both SFT and RL training. As shown in Table~\ref{tab:ablation_relative}, including OCR objectives yields consistent performance gains across benchmarks. 
This suggests that explicitly reinforcing low-level text recognition helps the model build stronger representations, which in turn improves long-context understanding ability.

\paragraph{Extreme Compression Exploration}

To further examine the potential of our approach, we explore more aggressive compression settings. 
We apply a configuration with an effective $8\times$ compression ratio during post-training, and evaluate the resulting model on MRCR with sequence lengths extended from 128k to 1024k. 
The results (Table \ref{tab:mrcr-128k}) show that \model successfully demonstrates the potential for $8\times$ effective context expansion, achieving performance on par with GLM-4-9B-Chat-1M and Qwen2.5-1M.  
This experiment highlights that our method can indeed be pushed to more extreme compression regimes while retaining performance, suggesting substantial headroom for extending usable context far beyond current limits, like a model that can deal with 4M, even 8M context tokens.


\section{Conclusion}

In this work, we present \model, an efficient long-context modeling framework that renders long texts into compact images and processes them with vision-language models. 
With continual pre-training, an LLM-driven genetic rendering search and targeted post-training, \model achieves 3–4× context compression while maintaining competitive performance with similar size leading LLMs such as Qwen3-8B. 
Extensive experiments further demonstrate substantial gains in inference speed and memory efficiency, and show that our method demonstrates cross-modal benefits, enhancing multimodal long-context tasks like document understanding. 
Our findings demonstrate that enhancing token information density constitutes a promising new paradigm for scaling long-context LLMs, orthogonal to existing attention-based approaches, and there remains great room for further exploration in depth.

\section*{Limitations and Future Work}

Despite the effectiveness of \model and its strong potential for broader applications, we want to discuss several limitations of the current work that are worth further exploration.  

\paragraph{Sensitivity to rendering parameters.}  
Our method relies on rendering textual inputs into images before processing.
We find that performance can be noticeably affected by rendering configurations such as resolution, font, and spacing. 
Although our search procedure allows us to identify a configuration that performs well on downstream tasks, how to make the model more robust across various rendering settings remains an open problem.

\paragraph{OCR-related challenges.}  
As discussed in the Ruler benchmark, UUID recognition remains particularly challenging for current VLMs, and even the strongest models (e.g., Gemini-2.5-Pro) often fail to reproduce them correctly. 
Such rare alphanumeric sequences frequently result in misordered or misclassified characters, which may stem from their distributional sparsity in training data or from architectural limitations of visual encoders. 
While these cases have little impact on most tasks, improving OCR fidelity could push the upper bound of our approach.

\paragraph{Task diversity.}  
The benchmarks in this work mainly focus on long-context understanding.
While these tasks provide a strong proof of concept, they do not fully capture the diversity of real-world applications, such as agentic or reasoning-heavy tasks. 
We also observe that, compared with textual models, the visual-text model tends to generalize less effectively across tasks.
Extending the scope of evaluation and training to a wider range of tasks will help better assess and improve the robustness and generality of our approach.

\paragraph{Future directions.}
Building upon the current study, several directions could further advance the proposed visual–text compression paradigm.
First, rather than using a fixed rendering strategy, one promising avenue is to train adaptive rendering models that condition on the task type or user query, producing tailored visualizations that balance compression and performance.
Second, enhancing the visual encoder's capability for fine-grained text recognition and alignment with language representations could improve robustness and transferability across tasks.
Third, improving the alignment between visual–text and purely textual models, for instance, through knowledge distillation or cross-modal supervision, could narrow the performance gap in generalization.
Fourth, our approach could be extended to broader applications, such as agent memory systems capable of managing long-term conversations or agentic contexts, and tasks that can leverage structured visual layouts for reasoning and retrieval.
From the perspective of context engineering, this method offers a new way to optimize how contextual information is represented and managed.
With further advances along this line, future models could go beyond the current limits of context length, effectively scaling from 1M to 10M input tokens.

\bibliography{custom}

\begin{thebibliography}{36}
\providecommand{\natexlab}[1]{#1}

\bibitem[{Bai et~al.(2025)Bai, Chen, Liu, Wang, Ge, Song, Dang, Wang, Wang, Tang et~al.}]{bai2025qwen2}
Shuai Bai, Keqin Chen, Xuejing Liu, Jialin Wang, Wenbin Ge, Sibo Song, Kai Dang, Peng Wang, Shijie Wang, Jun Tang, and 1 others. 2025.
\newblock Qwen2. 5-vl technical report.
\newblock \emph{arXiv preprint arXiv:2502.13923}.

\bibitem[{Bai et~al.(2024)Bai, Lv, Zhang, Lyu, Tang, Huang, Du, Liu, Zeng, Hou et~al.}]{bai2024longbench}
Yushi Bai, Xin Lv, Jiajie Zhang, Hongchang Lyu, Jiankai Tang, Zhidian Huang, Zhengxiao Du, Xiao Liu, Aohan Zeng, Lei Hou, and 1 others. 2024.
\newblock Longbench: A bilingual, multitask benchmark for long context understanding.
\newblock In \emph{Proceedings of the 62nd Annual Meeting of the Association for Computational Linguistics (Volume 1: Long Papers)}, pages 3119--3137.

\bibitem[{Beltagy et~al.(2020)Beltagy, Peters, and Cohan}]{longformer}
Iz~Beltagy, Matthew~E. Peters, and Arman Cohan. 2020.
\newblock Longformer: The long-document transformer.
\newblock In \emph{Proceedings of ACL}.

\bibitem[{Brown et~al.(2020)Brown, Mann, Ryder, Subbiah, Kaplan, Dhariwal, Neelakantan, Shyam, Sastry, Askell, Agarwal, Herbert-Voss, Krueger, Henighan, Child, Ramesh, Ziegler, Wu, Winter, Hesse, Chen, Sigler, Litwin, Gray, Chess, Clark, Berner, McCandlish, Radford, Sutskever, and Amodei}]{GPT3}
Tom~B. Brown, Benjamin Mann, Nick Ryder, Melanie Subbiah, Jared Kaplan, Prafulla Dhariwal, Arvind Neelakantan, Pranav Shyam, Girish Sastry, Amanda Askell, Sandhini Agarwal, Ariel Herbert-Voss, Gretchen Krueger, Tom Henighan, Rewon Child, Aditya Ramesh, Daniel~M. Ziegler, Jeffrey Wu, Clemens Winter, and 12 others. 2020.
\newblock Language models are few-shot learners.
\newblock In \emph{Proceedings of the 34th International Conference on Neural Information Processing Systems}, NIPS'20, Red Hook, NY, USA. Curran Associates Inc.

\bibitem[{Chen et~al.(2025{\natexlab{a}})Chen, Li, Gong, Jiang, Fei, Yang, Shan, Yu, Wang, Zhu et~al.}]{chen2025minimax}
Aili Chen, Aonian Li, Bangwei Gong, Binyang Jiang, Bo~Fei, Bo~Yang, Boji Shan, Changqing Yu, Chao Wang, Cheng Zhu, and 1 others. 2025{\natexlab{a}}.
\newblock Minimax-m1: Scaling test-time compute efficiently with lightning attention.
\newblock \emph{arXiv preprint arXiv:2506.13585}.

\bibitem[{Chen et~al.(2022)Chen, Wang, Changpinyo, Piergiovanni, Padlewski, Salz, Goodman, Grycner, Mustafa, Beyer et~al.}]{chen2022pali}
Xi~Chen, Xiao Wang, Soravit Changpinyo, Anthony~J Piergiovanni, Piotr Padlewski, Daniel Salz, Sebastian Goodman, Adam Grycner, Basil Mustafa, Lucas Beyer, and 1 others. 2022.
\newblock Pali: A jointly-scaled multilingual language-image model.
\newblock \emph{arXiv preprint arXiv:2209.06794}.

\bibitem[{Chen et~al.(2024)Chen, Qian, Tang, Lai, Liu, Han, and Jia}]{longlora}
Yukang Chen, Shengju Qian, Haotian Tang, Xin Lai, Zhijian Liu, Song Han, and Jiaya Jia. 2024.
\newblock Longlora: Efficient fine-tuning of long-context large language models.
\newblock In \emph{The International Conference on Learning Representations (ICLR)}.

\bibitem[{Chen et~al.(2025{\natexlab{b}})Chen, Wang et~al.}]{cope}
Yutao Chen, Yiren Wang, and 1 others. 2025{\natexlab{b}}.
\newblock Cope: Complex positional encoding for long context extrapolation.
\newblock \emph{arXiv preprint arXiv:2508.18308}.

\bibitem[{Chowdhery et~al.(2022)Chowdhery, Narang, Devlin, Bosma, Mishra, Roberts, Barham, Chung, Sutton, Gehrmann et~al.}]{chowdhery2022palm}
Aakanksha Chowdhery, Sharan Narang, Jacob Devlin, Maarten Bosma, Gaurav Mishra, Adam Roberts, Paul Barham, Hyung~Won Chung, Charles Sutton, Sebastian Gehrmann, and 1 others. 2022.
\newblock Palm: Scaling language modeling with pathways.
\newblock \emph{arXiv preprint arXiv:2204.02311}.

\bibitem[{Comanici et~al.(2025)Comanici, Bieber, Schaekermann, Pasupat, Sachdeva, Dhillon, Blistein, Ram, Zhang, Rosen et~al.}]{comanici2025gemini}
Gheorghe Comanici, Eric Bieber, Mike Schaekermann, Ice Pasupat, Noveen Sachdeva, Inderjit Dhillon, Marcel Blistein, Ori Ram, Dan Zhang, Evan Rosen, and 1 others. 2025.
\newblock Gemini 2.5: Pushing the frontier with advanced reasoning, multimodality, long context, and next generation agentic capabilities.
\newblock \emph{arXiv preprint arXiv:2507.06261}.

\bibitem[{GLM et~al.(2024)GLM, :, Zeng, Xu, Wang, Zhang, Yin, Rojas, Feng, Zhao, Lai, Yu, Wang, Sun, Zhang, Cheng, Gui, Tang, Zhang, Li, Zhao, Wu, Zhong, Liu, Huang, Zhang, Zheng, Lu, Duan, Zhang, Cao, Yang, Tam, Zhao, Liu, Xia, Zhang, Gu, Lv, Liu, Liu, Yang, Song, Zhang, An, Xu, Niu, Yang, Li, Bai, Dong, Qi, Wang, Yang, Du, Hou, and Wang}]{glm2024chatglm}
Team GLM, :, Aohan Zeng, Bin Xu, Bowen Wang, Chenhui Zhang, Da~Yin, Diego Rojas, Guanyu Feng, Hanlin Zhao, Hanyu Lai, Hao Yu, Hongning Wang, Jiadai Sun, Jiajie Zhang, Jiale Cheng, Jiayi Gui, Jie Tang, Jing Zhang, and 38 others. 2024.
\newblock \href {https://arxiv.org/abs/2406.12793} {Chatglm: A family of large language models from glm-130b to glm-4 all tools}.
\newblock \emph{Preprint}, arXiv:2406.12793.

\bibitem[{Hong et~al.(2024{\natexlab{a}})Hong, Wang, Ding, Yu, Lv, Wang, Cheng, Huang, Ji, Xue et~al.}]{hong2024cogvlm2}
Wenyi Hong, Weihan Wang, Ming Ding, Wenmeng Yu, Qingsong Lv, Yan Wang, Yean Cheng, Shiyu Huang, Junhui Ji, Zhao Xue, and 1 others. 2024{\natexlab{a}}.
\newblock Cogvlm2: Visual language models for image and video understanding.
\newblock \emph{arXiv preprint arXiv:2408.16500}.

\bibitem[{Hong et~al.(2024{\natexlab{b}})Hong, Wang, Lv, Xu, Yu, Ji, Wang, Wang, Dong, Ding et~al.}]{hong2024cogagent}
Wenyi Hong, Weihan Wang, Qingsong Lv, Jiazheng Xu, Wenmeng Yu, Junhui Ji, Yan Wang, Zihan Wang, Yuxiao Dong, Ming Ding, and 1 others. 2024{\natexlab{b}}.
\newblock Cogagent: A visual language model for gui agents.
\newblock In \emph{Proceedings of the IEEE/CVF Conference on Computer Vision and Pattern Recognition}, pages 14281--14290.

\bibitem[{Huang et~al.(2023)Huang, Xu, Lai, Jiang, Chen, Li, Yao, Ma, Yang, Chen et~al.}]{huang2023advancing}
Yunpeng Huang, Jingwei Xu, Junyu Lai, Zixu Jiang, Taolue Chen, Zenan Li, Yuan Yao, Xiaoxing Ma, Lijuan Yang, Hao Chen, and 1 others. 2023.
\newblock Advancing transformer architecture in long-context large language models: A comprehensive survey.
\newblock \emph{arXiv preprint arXiv:2311.12351}.

\bibitem[{Hurst et~al.(2024)Hurst, Lerer, Goucher, Perelman, Ramesh, Clark, Ostrow, Welihinda, Hayes, Radford et~al.}]{hurst2024gpt}
Aaron Hurst, Adam Lerer, Adam~P Goucher, Adam Perelman, Aditya Ramesh, Aidan Clark, AJ~Ostrow, Akila Welihinda, Alan Hayes, Alec Radford, and 1 others. 2024.
\newblock Gpt-4o system card.
\newblock \emph{arXiv preprint arXiv:2410.21276}.

\bibitem[{Laban et~al.(2024)Laban, Fabbri, Xiong, and Wu}]{laban2024summary}
Philippe Laban, Alexander~Richard Fabbri, Caiming Xiong, and Chien-Sheng Wu. 2024.
\newblock Summary of a haystack: A challenge to long-context llms and rag systems.
\newblock In \emph{Proceedings of the 2024 Conference on Empirical Methods in Natural Language Processing}, pages 9885--9903.

\bibitem[{Liu et~al.(2024{\natexlab{a}})Liu, Li, Li, Li, Zhang, Shen, and Lee}]{liu2024llavanext}
Haotian Liu, Chunyuan Li, Yuheng Li, Bo~Li, Yuanhan Zhang, Sheng Shen, and Yong~Jae Lee. 2024{\natexlab{a}}.
\newblock \href {https://llava-vl.github.io/blog/2024-01-30-llava-next/} {Llava-next: Improved reasoning, ocr, and world knowledge}.

\bibitem[{Liu et~al.(2023)Liu, Li, Wu, and Lee}]{liu2023visual}
Haotian Liu, Chunyuan Li, Qingyang Wu, and Yong~Jae Lee. 2023.
\newblock Visual instruction tuning.
\newblock \emph{Advances in neural information processing systems}, 36:34892--34916.

\bibitem[{Liu et~al.(2024{\natexlab{b}})Liu, Gao et~al.}]{prolong}
Yang Liu, Wei Gao, and 1 others. 2024{\natexlab{b}}.
\newblock Long context is not long at all: A prospector of long-dependency data for large language models.
\newblock \emph{arXiv preprint arXiv:2407.11234}.

\bibitem[{Peng et~al.(2023)Peng, Li et~al.}]{yarn}
Baolin Peng, Zhuohan Li, and 1 others. 2023.
\newblock Yarn: Efficient context window extension of large language models.
\newblock \emph{arXiv preprint arXiv:2309.00071}.

\bibitem[{Peng et~al.(2025)Peng, Zhang, Goldstein, Alcaide, Du, Hou, Lin, Liu, Lu, Merrill et~al.}]{peng2025rwkv}
Bo~Peng, Ruichong Zhang, Daniel Goldstein, Eric Alcaide, Xingjian Du, Haowen Hou, Jiaju Lin, Jiaxing Liu, Janna Lu, William Merrill, and 1 others. 2025.
\newblock Rwkv-7" goose" with expressive dynamic state evolution.
\newblock \emph{arXiv preprint arXiv:2503.14456}.

\bibitem[{Press et~al.(2021)Press, Smith, and Levy}]{alibi}
Ofir Press, Noah~A. Smith, and Mike Levy. 2021.
\newblock Train short, test long: Attention with linear biases enables input length extrapolation.
\newblock \emph{arXiv preprint arXiv:2108.12409}.

\bibitem[{Su et~al.(2021)Su, Lu, Pan, Wen, and Liu}]{rope}
Jianlin Su, Yu~Lu, Shengfeng Pan, Bo~Wen, and Yunfeng Liu. 2021.
\newblock Roformer: Enhanced transformer with rotary position embedding.
\newblock \emph{arXiv preprint arXiv:2104.09864}.

\bibitem[{Sun et~al.(2022)Sun, Cheng, He et~al.}]{xpos}
Zhen Sun, Peng Cheng, Wei He, and 1 others. 2022.
\newblock Xpos: Improving position interpolation with extrapolation.
\newblock \emph{arXiv preprint arXiv:2212.10554}.

\bibitem[{Touvron et~al.(2023)Touvron, Lavril, Izacard, Martinet, Lachaux, Lacroix, Rozi{\`e}re, Goyal, Hambro, Azhar et~al.}]{touvron2023llama}
Hugo Touvron, Thibaut Lavril, Gautier Izacard, Xavier Martinet, Marie-Anne Lachaux, Timoth{\'e}e Lacroix, Baptiste Rozi{\`e}re, Naman Goyal, Eric Hambro, Faisal Azhar, and 1 others. 2023.
\newblock Llama: Open and efficient foundation language models.
\newblock \emph{arXiv preprint arXiv:2302.13971}.

\bibitem[{Vodrahalli et~al.(2024)Vodrahalli, Ontanon, Tripuraneni, Xu, Jain, Shivanna, Hui, Dikkala, Kazemi, Fatemi et~al.}]{vodrahalli2024michelangelo}
Kiran Vodrahalli, Santiago Ontanon, Nilesh Tripuraneni, Kelvin Xu, Sanil Jain, Rakesh Shivanna, Jeffrey Hui, Nishanth Dikkala, Mehran Kazemi, Bahare Fatemi, and 1 others. 2024.
\newblock Michelangelo: Long context evaluations beyond haystacks via latent structure queries.
\newblock \emph{arXiv preprint arXiv:2409.12640}.

\bibitem[{Wang et~al.(2024{\natexlab{a}})Wang, Lv, Yu, Hong, Qi, Wang, Ji, Yang, Zhao, XiXuan et~al.}]{wang2024cogvlm}
Weihan Wang, Qingsong Lv, Wenmeng Yu, Wenyi Hong, Ji~Qi, Yan Wang, Junhui Ji, Zhuoyi Yang, Lei Zhao, Song XiXuan, and 1 others. 2024{\natexlab{a}}.
\newblock Cogvlm: Visual expert for pretrained language models.
\newblock \emph{Advances in Neural Information Processing Systems}, 37:121475--121499.

\bibitem[{Wang et~al.(2024{\natexlab{b}})Wang, Yang et~al.}]{longrecipe}
Yizhi Wang, Fan Yang, and 1 others. 2024{\natexlab{b}}.
\newblock Longrecipe: Recipe for efficient long context generalization in large language models.
\newblock \emph{arXiv preprint arXiv:2406.12345}.

\bibitem[{Wu et~al.(2024)Wu, Gu, Feng, Zhong, Xu, Yang, Liu, and Qin}]{wuextending}
Yingsheng Wu, Yuxuan Gu, Xiaocheng Feng, Weihong Zhong, Dongliang Xu, Qing Yang, Hongtao Liu, and Bing Qin. 2024.
\newblock Extending context window of large language models from a distributional perspective.
\newblock \emph{arXiv preprint arXiv:2410.01490}.

\bibitem[{Yang et~al.(2025)Yang, Li, Yang, Zhang, Hui, Zheng, Yu, Gao, Huang, Lv et~al.}]{yang2025qwen3}
An~Yang, Anfeng Li, Baosong Yang, Beichen Zhang, Binyuan Hui, Bo~Zheng, Bowen Yu, Chang Gao, Chengen Huang, Chenxu Lv, and 1 others. 2025.
\newblock Qwen3 technical report.
\newblock \emph{arXiv preprint arXiv:2505.09388}.

\bibitem[{Yang et~al.(2024)Yang, Wang, Shen, Panda, and Kim}]{yang2024gated}
Songlin Yang, Bailin Wang, Yikang Shen, Rameswar Panda, and Yoon Kim. 2024.
\newblock Gated linear attention transformers with hardware-efficient training.
\newblock In \emph{International Conference on Machine Learning}, pages 56501--56523. PMLR.

\bibitem[{Yang et~al.(2016)Yang, Yang, Dyer, He, Smola, and Hovy}]{han}
Zichao Yang, Diyi Yang, Chris Dyer, Xiaodong He, Alex Smola, and Eduard Hovy. 2016.
\newblock Hierarchical attention networks for document classification.
\newblock In \emph{Proceedings of NAACL}.

\bibitem[{Yu et~al.(2025{\natexlab{a}})Yu, Chen, Feng, Chen, Dai, Yu, Zhang, Ma, Liu, Wang et~al.}]{yu2025memagent}
Hongli Yu, Tinghong Chen, Jiangtao Feng, Jiangjie Chen, Weinan Dai, Qiying Yu, Ya-Qin Zhang, Wei-Ying Ma, Jingjing Liu, Mingxuan Wang, and 1 others. 2025{\natexlab{a}}.
\newblock Memagent: Reshaping long-context llm with multi-conv rl-based memory agent.
\newblock \emph{arXiv preprint arXiv:2507.02259}.

\bibitem[{Yu et~al.(2025{\natexlab{b}})Yu, Zhang, Zhu, Yuan, Zuo, Yue, Dai, Fan, Liu, Liu et~al.}]{yu2025dapo}
Qiying Yu, Zheng Zhang, Ruofei Zhu, Yufeng Yuan, Xiaochen Zuo, Yu~Yue, Weinan Dai, Tiantian Fan, Gaohong Liu, Lingjun Liu, and 1 others. 2025{\natexlab{b}}.
\newblock Dapo: An open-source llm reinforcement learning system at scale.
\newblock \emph{arXiv preprint arXiv:2503.14476}.

\bibitem[{Zhang et~al.(2024)Zhang, Li et~al.}]{longalign}
Tianle Zhang, Zhuohan Li, and 1 others. 2024.
\newblock Longalign: Instruction-tuning long-context llms.
\newblock \emph{arXiv preprint arXiv:2401.10968}.

\bibitem[{Zhu et~al.(2024)Zhu, Wang et~al.}]{dape}
Wei Zhu, Ziheng Wang, and 1 others. 2024.
\newblock Data-adaptive positional encoding for length generalization.
\newblock \emph{NeurIPS}.

\end{thebibliography}

\appendix

\section{Rendering Parameters}

\begin{figure*}[h!]
    \centering
    \includegraphics[width=0.8\textwidth]{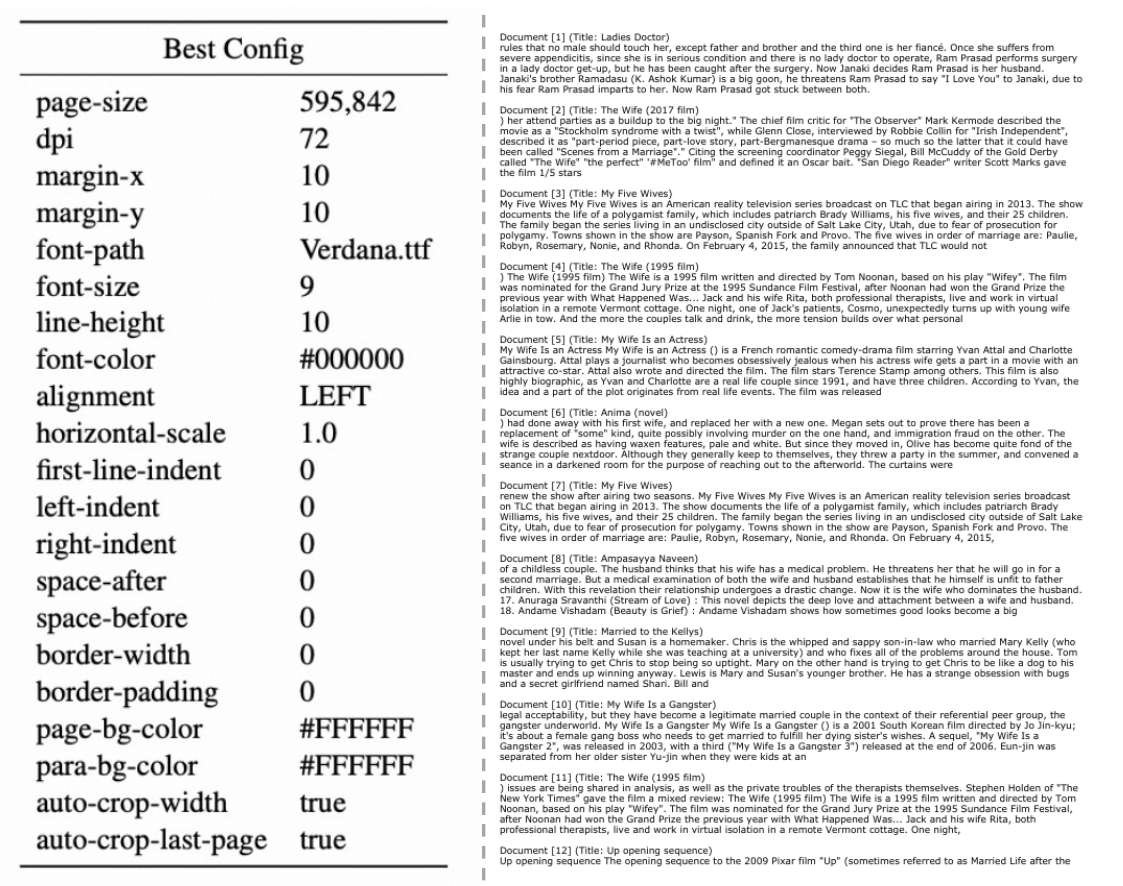}
    \caption{The optimal parameter setting. The left column lists the values for page layout, font, and spacing, while the right column provides an example of the rendered text.}
    \label{fig:best_config}
\end{figure*}

Our rendering parameters are detailed in Table \ref{tab:rendering-params}. The best result, obtained through our LLM-driven genetic search, is presented in Figure~\ref{fig:best_config}, which shows the detailed configuration and its corresponding rendering.

\begin{table}[htbp]
\centering
\small
\setlength{\tabcolsep}{5pt}
\begin{tabular}{p{0.28\linewidth} p{0.62\linewidth}}
\toprule
\textbf{Factor} & \textbf{Specification / Sampling Strategy} \\
\midrule
\texttt{dpi} & Mixture of sets: \emph{lowest} (45--59), \emph{low} (60--71), \emph{medium} (72--119), \emph{normal} (\{72,80,96,100,110,120,144,150,300\}), \emph{high} (over 300); favor normal/medium with small probability spikes to extremes. \\
\texttt{page\_size} & (i) Fixed paper sizes (A4, Letter, Legal, A5, B5, A3, B4, Tabloid) with priors; (ii) common aspect ratios (e.g., 1.414, 1.333, 1.5, 1.778); (iii) fully random aspect via piecewise distribution (narrow $\rightarrow$ tall). \\
\texttt{font\_family} & Pooled and deduplicated families across serif/sans/mono/pixel; italics sampled by filename heuristics (suffixes, \texttt{italic}/\texttt{oblique}). \\
\texttt{font\_size} & $\{7, 7.5, 8, 9, 9.5, 10, 11, 12, 14\}$; \texttt{line\_height} tied as \texttt{font\_size} + $\{0,\dots,3\}$. \\
\texttt{alignment} & \textsc{Left}/\textsc{Justify} (dominant) with small-prob.\ \textsc{Right}/\textsc{Center}. \\
\texttt{margins} & Three patterns: all-equal; vertical-larger; horizontal-larger; values in 10--40pt ranges. \\
\texttt{indent} & Modes: none; first-line indent ($\approx$1--2.5\,em); block/hanging with left/right indents. \\
\texttt{spacing} & \texttt{space-before}/\texttt{space-after} use a multi-mode prior (none, small, large). \\
\texttt{h\_scale} & Horizontal glyph scaling (0.75--1.0) with decaying probabilities. \\
\texttt{colors} & Page/background/font palettes for light/dark themes; document/web/code styles inherit coherent triplets (page, paragraph, font). \\
\texttt{borders} & Optional paragraph borders with width/padding; disabled by default. \\
\texttt{newline\_markup} & With small probability, explicit markers (e.g., \verb|\n|, tags, or tokens) inserted to preserve structure. \\
\texttt{auto\_crop} & Optional white-margin cropping and last-page trimming. \\
\bottomrule
\end{tabular}
\caption{Controllable factors in the rendering pipeline and their sampling strategies.
The mixture design yields broad yet realistic typography/layout coverage and tunable compression $\rho(\boldsymbol{\theta})$.}
\label{tab:rendering-params}
\end{table}

\begin{table}[htbp]
\centering
\label{tab:mrcr-2needle}
\resizebox{0.95\linewidth}{!}{%
\begin{tabular}{lcccccc}
\toprule
\multirow{2}{*}{Model} & \multicolumn{6}{c}{2 Needle} \\
\cmidrule(lr){2-7}
& 0k-8k & 8k-16k & 16k-32k & 32k-64k & 64k-128k & Avg \\
\midrule
\textcolor{gray}{GPT-4.1} & \textcolor{gray}{83} & \textcolor{gray}{72} & \textcolor{gray}{67} & \textcolor{gray}{62} & \textcolor{gray}{59} & \textcolor{gray}{68.6} \\
\hdashline
\addlinespace
LLaMA-3.1-8B-Instruct  & \underline{54.27} & \textbf{53.21} & \textbf{51.05} & \underline{29.81} & \underline{24.98} & \underline{42.66} \\
Qwen3-8B     & \textbf{58.95} & 41.18 & 36.18 & 24.99 & 20.89 & 36.44 \\
GLM-4-9B-Chat-1M   & 39.77 & 15.87 & 18.42 & 18.63 & 18.42 & 22.22 \\
Qwen2.5-7B-Instruct-1M   & 45.92 & \underline{51.07} & \underline{46.97} & \textbf{34.67} & \textbf{37.57} & \textbf{43.24} \\
\midrule
Glyph        & 41.51 & 40.78 & 39.58 & 29.67 & 22.41 & 34.85 \\
\bottomrule
\end{tabular}%
}
\caption{Performance of various models on the MRCR task (\%) with the 2 Needle setting across different context length intervals (0k–8k, 8k–16k, 16k–32k, 32k–64k, 64k–128k) and the average score.}
\end{table}

\begin{table*}[htbp]
\centering
\resizebox{0.95\linewidth}{!}{%
\begin{tabular}{@{}l ccccccccc@{}}  
\toprule
\multirow{2}{*}{\textbf{Model}} 
& \multicolumn{2}{c}{\textbf{Single-Doc QA}} 
& \multicolumn{2}{c}{\textbf{Multi-Doc QA}} 
& \multicolumn{2}{c}{\textbf{Summarization}} 
& \multicolumn{2}{c}{\textbf{Few-shot}} 
& \multicolumn{1}{c}{\textbf{Synthetic}} \\  
\cmidrule(lr){2-3} \cmidrule(lr){4-5} \cmidrule(lr){6-7} \cmidrule(lr){8-9} \cmidrule(lr){10-10}
& \textbf{QA Zh} & \textbf{QA En} & \textbf{Mus} & \textbf{Dur} & \textbf{News} & \textbf{VcSum} & \textbf{Sam} & \textbf{Lsht} & \textbf{Pa C} \\
\midrule
\textcolor{gray}{GPT-4.1} & \textcolor{gray}{63.90} & \textcolor{gray}{51.27} & \textcolor{gray}{55.63} & \textcolor{gray}{24.58} & \textcolor{gray}{23.70} & \textcolor{gray}{14.66} & \textcolor{gray}{41.25} & \textcolor{gray}{50.00} & \textcolor{gray}{26.5} \\
\hdashline 
\addlinespace 
LLaMA-3.1-8B-Instruct   & 62.20 & \textbf{54.98} & 31.61 & \textbf{33.75} & \textbf{24.21} & \textbf{16.23} &  7.61 &  0.00 &  7.13 \\
Qwen3-8B      & 60.98 & 49.78 & \underline{45.54} & 16.69 & 18.55 & 12.08 & \underline{36.47} & 42.00 & \underline{12.81} \\
GLM-4-9B-Chat-1M    & \textbf{63.17} & 52.88 & 39.14 & \underline{28.27} & \underline{23.90} & \underline{16.21} & 36.15 & \textbf{47.38} &  2.39 \\
Qwen2.5-7B-Instruct-1M    & \underline{62.98} & \underline{53.62} & 34.72 & 21.85 & 21.02 & 12.20 & \textbf{39.17} & 28.68 &  3.50 \\
\midrule
\model         & 37.23 & 45.89 & \textbf{56.18} & 26.87 & 21.52 & 12.43 & 32.49 & \underline{44.43} & \textbf{30.50} \\
\bottomrule
\end{tabular}%
}
\caption{The rest of the results on LongBench benchmark (\%), which encompasses Single-Document QA, Multi-Document QA, Summarization, Few-shot Learning, and Synthetic task.}
\label{tab:longbench_subtasks}
\end{table*}

\section{Implementation Details} \label{appendix: implementatio} 

\paragraph{Training Details}
For continual pre-training of the 9B long-context backbone, the model is initialized from the released GLM-4.1V-9B-Base checkpoint and trained on a diverse mixture of rendered long-context data and vision-language corpora (e.g., OCR task) within 128k context length.
The training uses a global batch size of 170 and a learning rate of 2e-6 with cosine decay for around 4000 steps.

For the rendering search, we run for 5 times with 200 steps in each round, to find the optimal configuration that maximizes the compression ratio while maintaining good performance.

After this, we conduct further SFT and RL training. For SFT, we train for 1.5k steps with a batch size of 32. The Adam optimizer ($\beta_1=0.9$, $\beta_2=0.95$) is used with cosine decay and 160 warm-up steps, where the learning rate decays from 5e-6 to 2e-6. 
For reinforcement learning, we adopt the GRPO algorithm. Each training group samples 16 candidate responses, and degenerate samples with all-zero or all-one rewards are discarded. 
We apply the clip-higher trick from DAPO \cite{yu2025dapo} with $\epsilon_l$ being 0.2 and $\epsilon_h$ being 0.28. Training runs for 500 iterations with a batch size of 32. We also use the Adam optimizer with a constant learning rate of 1e-6.

\paragraph{Baselines.}
We compare \model with leading open-sourced LLMs of similar size:
\begin{itemize}
    \item \textbf{Qwen3-8B} achieves state-of-the-art performance across reasoning and a wide range of tasks.
    \item \textbf{Qwen2.5-7B-Instruct-1M} excels at long-context understanding and achieves strong performance across diverse benchmarks.
    \item \textbf{LLaMA-3.1-8B-Instruct} is a widely used model with strong instruction-following and multilingual capabilities.
    \item \textbf{GLM-4-9B-Chat-1M} delivers powerful long-context tasks and overall high performance.
\end{itemize}

\paragraph{Backbone Model.} Our method relies on a strong VLM to process long-context tasks. Considering the impressive performance of GLM-4.1V-9B, especially in OCR and long document tasks, we have chosen GLM-4.1V-9B-Base as our backbone model.

\paragraph{Evaluation Benchmarks.}
To conduct a comprehensive analysis of the long-context performance, we have adopted three popular benchmarks, including LongBench, MRCR, and Ruler. LongBench consists of 21 datasets in total in 6 categories, covering diverse long-context tasks. MRCR is a task proposed by \citet{vodrahalli2024michelangelo}. We use the OpenAI version, which consists of multi-turn conversations about writing, asking models to recall one of the contexts in dialogue history. Ruler is a widely used synthetic benchmark with 11 NIAH tasks.
To validate cross-model benefits, we choose the MMLongBench-Doc, which involves 130 lengthy PDF with diverse layout and images, and 1062 questions.

\paragraph{Efficiency Evaluation Setting.}
For training, we focus on SFT since RL involves rollout time, making it difficult to compare fairly. Moreover, running RL at very long context lengths (e.g., 64k or beyond) requires excessive memory resources. This again highlights the advantage of our approach: through compression, we can conduct RL training at 32k context length while effectively covering over 100k tokens of raw text input, where RL is prohibitively difficult for LLMs due to memory and computation demands.
For SFT, we measure per-sample training time under the same number of training data using 8$\times$80G H100 GPUs.
For inference, we deploy both models on a single 80G H100 and measure efficiency along two axes: (i) prefill latency at batch size 1, and (ii) per-sample inference time at the maximum feasible batch size with output length set to 256 tokens.
We omit the KV cache testing, because KV cache scales linearly with the sequence length, and compression translates almost directly into savings of about 67\% memory usage. 


\end{document}